\documentclass{article} 
\usepackage{iclr2027_conference,times}

\usepackage{amsmath,amsfonts,bm}

\def\eqref#1{equation~\ref{#1}}

\def\1{\bm{1}}

\DeclareMathAlphabet{\mathsfit}{\encodingdefault}{\sfdefault}{m}{sl}
\SetMathAlphabet{\mathsfit}{bold}{\encodingdefault}{\sfdefault}{bx}{n}

\usepackage{hyperref}
\usepackage{url}

\usepackage{booktabs}
\usepackage{graphicx}
\usepackage{multirow}
\usepackage{makecell}
\usepackage{tabularx}
\usepackage{algorithm}
\usepackage{algpseudocode}
\usepackage{tcolorbox}
\usepackage{listings}
\usepackage{subcaption}
\usepackage{pifont}
\usepackage{array}
\usepackage{enumitem}
\tcbuselibrary{skins,breakable}

\usepackage{xcolor}

\definecolor{jsonbg}{RGB}{248,248,248}
\definecolor{jsonkey}{RGB}{30,80,160}

\lstdefinestyle{json}{
    basicstyle=\ttfamily\small,
    backgroundcolor=\color{jsonbg},
    frame=single,
    rulecolor=\color{black!20},
    breaklines=true,
    breakatwhitespace=false,
    columns=fullflexible,
    keepspaces=true,
    showstringspaces=false,
    xleftmargin=4pt,
    xrightmargin=4pt,
    aboveskip=4pt,
    belowskip=4pt
}

\newcommand{\framework}{\textsc{SimTrace}}

\title{\framework{}: Grounded Multimodal User Trajectories Generation for Online User Modeling}

\author{
 \textbf{Yunan Lu \thanks{~~This work was done as part of internship at Shopify.} \ \textsuperscript{1,2}},
 \textbf{Shuang Xie \textsuperscript{1}},
 \textbf{Meghna Allamudi \textsuperscript{2}},
 \textbf{Mingyu Zhao \textsuperscript{1}},
 \textbf{Han Li \textsuperscript{1}}, \\
 \ \textbf{Lingyun Wang \textsuperscript{1}},
 \textbf{Zhou Yu \textsuperscript{2}}
\\
 \textsuperscript{1}Shopify \space
 \textsuperscript{2}Columbia University
\\
   \textbf{Correspondence:} 
   \href{mailto:email@domain}{yl4021@columbia.edu}, 
   \href{mailto:email@domain}{lingyun.wang@shopify.com}
}

\iclrfinalcopy 
\begin{document}

\maketitle

\begin{abstract}
Virtual clients offer a cost-effective approach to support applications such as A/B testing, recommender system development, and interface evaluation. However, building them requires access to large-scale, semantically faithful, fine-grained online user trajectories. These data are difficult to obtain because proprietary logs are subject to privacy restrictions and small businesses often lack sufficient traffic. Consequently, existing public datasets either abstract away fine-grained user interaction details or preserve rich context but remain platform-specific and small-scale.
To address this gap, we propose \framework{}, a framework that generates faithful, fine-grained synthetic multimodal clickstreams through a computer-use client agent that is grounded in real user trajectories and the given web environment. \framework{} anonymizes real interactions and constructs a simulated twin of the given web environment, then uses both to generate synthetic interaction trajectories. Each action is paired with its corresponding web observations and user context, yielding a shareable alternative to confidential logs for developing computer-use agent-style virtual clients.
We apply \framework{} to an e-commerce setting and evaluate both its fidelity and downstream utility.  \framework{} outperforms competing baselines on 7 out of 8 fidelity metrics. Models trained on synthetic data achieve performance comparable to those trained on real data on downstream tasks such as purchase prediction and recommendation. For next action prediction task, augmenting real data with synthetic data further improves accuracy by 11.0\% relative to training on real data alone. We release \framework{} as an open-source package to facilitate research on online user behavior modeling.


\end{abstract}

\section{Introduction}
\label{sec:introduction}

LLMs are increasingly used to model online user behavior in web environments to support a broad range of applications, including A/B testing, recommender systems, usability studies, market research, and the evaluation of interactive LLM-based systems \citep{li_simgym_2026, lu_vista_2026, chen_recusersim_2025, sun_llm_2025, wang_large_2026}. Many of these applications require models not only to predict final outcomes but also to reproduce the step-by-step interactions leading to those outcomes. Recording these interactions as trajectories allows researchers and practitioners to inspect how user decisions unfold. This provides an interpretable record of the decision-making process. Therefore, fine-grained trajectories that faithfully pair each action with its corresponding web observation are essential for developing computer-user virtual clients for these applications.

However, existing public datasets generally either abstract away fine-grained user interaction details or preserve rich context but remain limited in scale. Large-scale datasets \citep{requena_shopper_2020, ben-shimon_recsys_2015, zhang_large_2015}, commonly used for recommendation and user behavior prediction, typically represent sessions as sequences of item identifiers or event types, omitting the interface observations and semantic context needed to train computer-use agents to reproduce users' interactions. Context-rich datasets \citep{wang_opera_2026, sun_llm_2025} collected through user studies, by contrast, preserve detailed interactions, but they are costly to construct, limited in scale, and typically confined to a single platform or environment, restricting their reuse in other settings. Organizations can instead collect interaction logs from their own environments, but this requires sufficient user traffic, and the resulting data are often proprietary and constrained by privacy regulations \citep{voigt_eu_2017}. Consequently, existing data sources rarely provide semantically faithful, environment-grounded trajectories at scale.

LLM-based user simulation provides a promising solution for generating context-rich trajectories at scale. Existing studies infer personas and intents from user activity logs or surveys and use persona-based LLM simulators to generate synthetic trajectories \citep{sun_llm_2025, li_simgym_2026}. However, these methods typically prioritize plausible task completion. Their generated step-by-step actions still differ substantially from those observed in real user sessions, as reported by \citet{zhang_shop-r1_2026}.

To address these limitations, we introduce \framework{}, a framework that uses a computer-use client agent to generate synthetic trajectories conditioned on real user trajectories and a given environment to improve the fidelity of step-by-step interactions in user modeling tasks.
Through empirical evaluation, we demonstrate that synthetic data generated by \framework{} outperform competing baselines \citep{li_simgym_2026, uxagent} on 7 out of 8 fidelity metrics.
Moreover, models trained on synthetic data perform comparably to those trained on real data across multiple downstream tasks. Augmenting real data with synthetic data further improves next action prediction accuracy by 11\%. Together, these results show that \framework{} can alleviate data scarcity and privacy constraints without compromising behavioral fidelity or downstream utility. We release \framework{} as an open-source package with modules for data collection, synthetic data generation, and user-model training, together with a synthetic multimodal computer-use trajectory dataset spanning five e-commerce environments. The package enables researchers to construct synthetic datasets for additional environments, facilitating the growth of shared resources for user behavior research. \footnote{The code is available at \url{https://github.com/agentic-foundation-modeling-research/SimTrace}.}

\section{\framework{}}
\label{sec:method}

\begin{figure}[t]
\begin{center}
\includegraphics[scale=0.50]{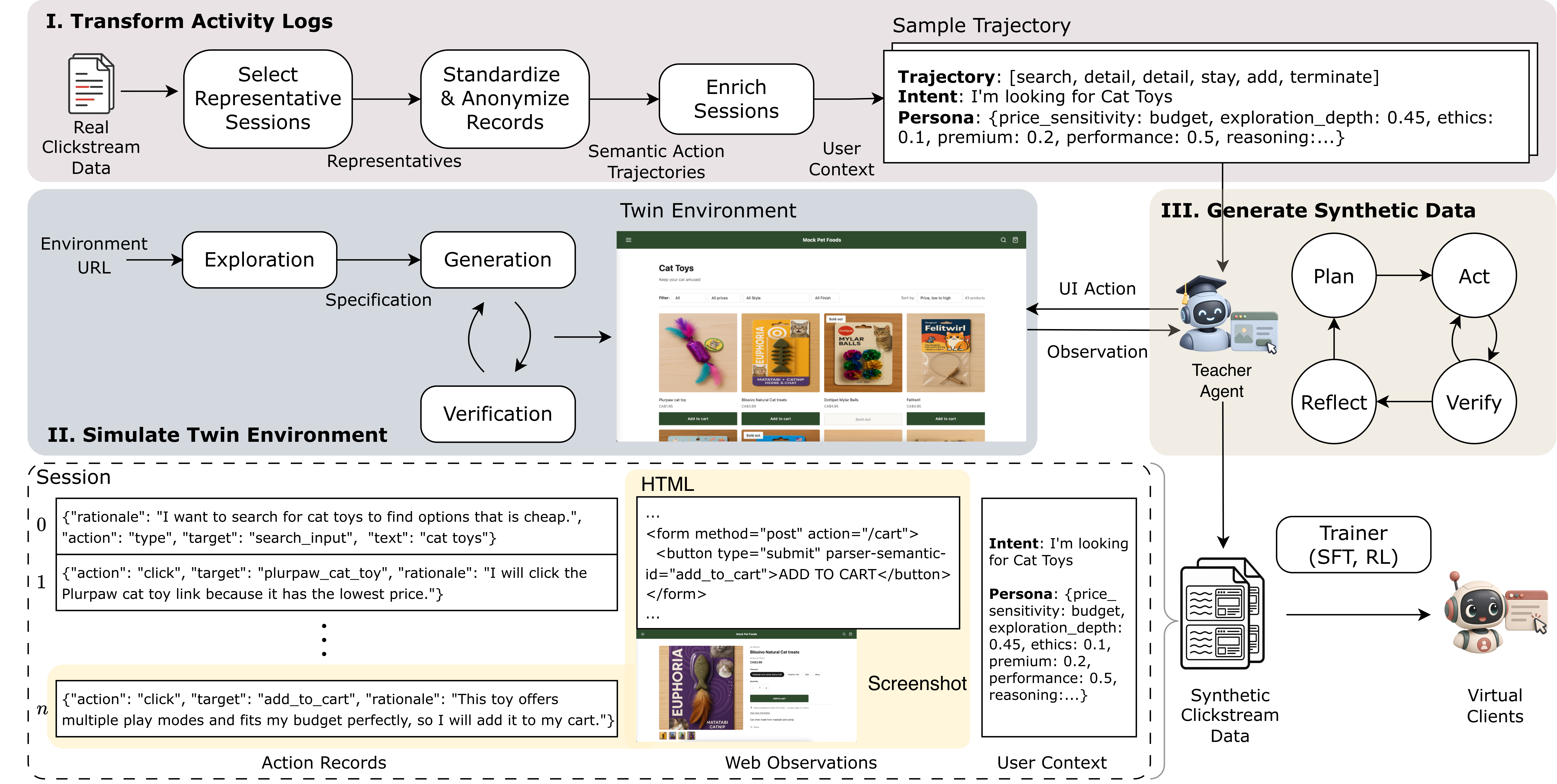}
\end{center}
\vspace{-0.6em}
\caption{Overview of \framework{} and the resulting dataset. \framework{} transforms real activity logs into representative, anonymized behavior trajectories, constructs a twin sandbox environment of the given website, and uses LLM-based teacher agent to generate synthetic interaction sessions. Each resulting session contains user context and a sequence of timestamped steps, where each step pairs a structured action record with the corresponding webpage HTML and screenshot. See Appendix \ref{app:data} for a complete data example.}
\label{fig:main}
\vspace{-1.2em}
\end{figure}

We introduce \framework{} as a scalable framework for generating semantically faithful multimodal clickstreams to support user behavior modeling.
As illustrated in Figure~\ref{fig:main}, \framework{} comprises three steps. 
First, it transforms noisy and private activity logs into abstract, anonymized behavioral trajectories (Section~\ref{sec:method:anon}). However, these abstract trajectories alone lack the fine-grained web observations required for user behavior modeling. To recover this interaction context in a controllable environment, \framework{} constructs a simulated twin grounded in the given web environment using a website-agnostic approach (Section~\ref{sec:method:env}). Finally, a teacher agent uses the anonymized trajectories as behavioral guidance and navigates the twin environment to generate fine-grained interactions paired with corresponding web observations and user context (Section~\ref{sec:method:gen}).

\subsection{Transform Activity Logs}
\label{sec:method:anon}
Activity logs record fine-grained interactions, including clicks, keystrokes, navigation paths, and time spent on each page \citep{bucklin_click_2009}. For environments without an existing logging pipeline, our package integrates with commonly used platforms, including Microsoft Clarity~\citep{microsoft_clarity} and PostHog~\citep{posthog_product_analytics}, to facilitate activity-log collection.

\vspace{-0.6em}
\paragraph{Select Representative Sessions.}
To obtain a compact yet representative set of user interactions, we select sessions that maximize the diversity of the given environment context and user behavior.
In the e-commerce setting, the environment context is represented by the range of products encountered, while user behavior is captured by three conversion-funnel outcomes \citep{Mittal-user-behavior}: Browse-only, Cart Abandonment, and Checkout. These outcomes are defined as follows: \emph{Browse-only} means that a user views products without adding any product to the cart or proceeding to checkout; \emph{Cart Abandonment} means that a user adds at least one product to the cart but does not proceed to checkout; and \emph{Checkout} means that a user proceeds to checkout.


Based on these representations, our objective is to select a subset of sessions that maximizes product coverage while maintaining a balanced distribution across the three outcomes. To solve this problem, we use a greedy algorithm as an efficient method that directly captures our selection criteria. The algorithm cycles through the three outcomes and selects the eligible session with the largest marginal coverage of previously unseen products. This iterative procedure terminates when it reaches the specified sample-size budget or a configured product-coverage threshold, such as 95\%. Appendix \ref{app:session-selection} provides implementation details and the pseudocode. 

\vspace{-0.6em}
\paragraph{Standardize and Anonymize Records.}

To provide a common interface between heterogeneous logging tools and downstream models, we standardize raw events using a semantic-action taxonomy. This standardization is necessary because logging tools differ in their event schemas and levels of granularity. Some tools record only high-level actions that trigger page transitions, whereas others capture low-level events such as individual mouse movements. These tool-specific and overly granular events can introduce noise and make activity logs difficult to compare across tools or use for behavior modeling. We therefore design a taxonomy that maps raw events to semantic actions in two categories: \emph{domain-specific actions}, which capture user goals and task-state changes specific to a particular domain, and \emph{domain-independent actions}, which capture navigation behavior that transfers across environments. For our e-commerce instantiation, we build on and extend the taxonomy introduced by \citet{requena_shopper_2020}, following the mutually exclusive and collectively exhaustive (MECE) principle \citep{mece_principle}. Table~\ref{tab:action_taxonomy} presents the resulting taxonomy.

After standardization, we further anonymize user, store, session, and product identifiers using keyed HMAC-SHA-256 hashes \citep{hmac1981}. We also abstract product information into a product category and a price bucket. Each transformed record contains \texttt{store\_id}, \texttt{session\_id}, \texttt{user\_id}, \texttt{timestamp}, \texttt{semantic\_action}, \texttt{product\_hash}, \texttt{product\_category}, and \texttt{price\_bucket}.

\vspace{-0.6em}
\paragraph{Enrich Sessions.}
Standardization and anonymization operate at the record level, but session-wide patterns can reveal latent user intent and preferences valuable for behavior modeling. We represent this session-level user context through \textbf{intent} and \textbf{persona} inferred from interaction history. For our e-commerce instantiation, we construct the persona following SimGym \citep{li_simgym_2026}. The persona comprises five dimensions: \emph{price sensitivity}, reflecting preferences across budget, mid-range, and premium products; \emph{exploration depth}, reflecting the extent of search and browsing; and \emph{premium focus}, \emph{performance focus}, and \emph{ethics focus}, capturing attention to luxury, durability, and ethical sourcing, respectively. See Appendix~\ref{app:persona_example}. The final transformed log represents each session through its semantic-action trajectory, inferred intent, and persona, as illustrated in Figure~\ref{fig:main}.

\begin{table}[t]
\centering
\fontsize{8.5pt}{8pt}\selectfont
\setlength{\tabcolsep}{6pt}
\caption{Taxonomy for transforming raw clickstream events in e-commerce domain.}
\label{tab:action_taxonomy}
\begin{tabular}{p{0.12\linewidth} p{0.17\linewidth} p{0.60\linewidth}}
\toprule
\textbf{Category} & \textbf{Semantic Action} & \textbf{Definition} \\
\midrule
& \texttt{search} & The user enters a query in the search bar. \\
\textbf{Domain-} & \texttt{detail} & The user visits a product page or interacts with product attributes. \\
\textbf{specific} & \texttt{add} & The user adds a product to the cart. \\
\textbf{Actions}& \texttt{remove} & The user removes a product from the cart. \\
& \texttt{checkout} & The user proceeds to purchase using \emph{Buy Now} or \emph{Checkout}. \\
\midrule
& \texttt{goto} & The user navigates to a new non-product-detail page, e.g. applying a filter, changing the sort order, or following a navigation link. \\
\textbf{Domain-} & \texttt{back} & The user returns to the previous page. \\
\textbf{independent Actions}& \texttt{stay} & The user remains on the same page while a new event is recorded, typically because of a same-page interaction, such as scrolling. \\
& \texttt{terminate} & The label is appended as the final action of every session. \\
\bottomrule
\end{tabular}
\vspace{-1em}
\end{table}

\vspace{-0.6em}
\subsection{Simulate Twin Environment}
\label{sec:method:env}

To provide web observations of corresponding actions, fine-grained trajectories need to be generated through interactions with an environment. To support controllable generation when direct interaction with the original environment is infeasible, such as when sensitive content must be masked, we provide a website-agnostic approach for constructing a simulated sandbox of a given environment.

Inspired by ShopGym~\citep{savadikar_shopgym_2026}, we autonomously construct the simulated sandbox through three modules: \textbf{exploration}, \textbf{generation}, and \textbf{verification}. During exploration, a powerful coding agent (e.g. \texttt{Pi} harness~\citep{pi} with \texttt{GPT-5.5}) follows the task lists we designed to inspect the original environment and collect browser evidence, including page screenshots. These observations are consolidated into specification capturing the visual theme, navigation structure, and supported interactions. The generation module, which is driven by the same agent, then transforms this specification into source code for a runnable sandbox while replacing sensitive content with synthetic alternatives. See Appendix \ref{app:env-gen} for implementation and cost details.
To ensure that the simulated environment preserves similar user experience, including user-observed content and user actions, we introduce two verifiers: a \emph{visual-fidelity verifier} and an \emph{action-replay verifier}. We modify the simulated environment iteratively until all the verifier feedback is resolved.

The \emph{visual-fidelity verifier} compares screenshots of the generated sandbox against reference screenshots collected during exploration. An LLM-based evaluator scores each screenshot from 0 to 10 along five dimensions: layout, color theme, component correspondence, content density, and language consistency. The sandbox passes visual verification only if the average score across these dimensions meets a configured threshold. Otherwise, the verifier returns concrete discrepancies and repair instructions for the next code-generation iteration.

The \emph{action-replay verifier} evaluates whether observed user trajectories remain executable in the simulated environment. It samples trajectories from real clickstreams and replays them in the simulated environment using Playwright browser automation. For example, a \texttt{detail} action passes if the associated product path returns a non-error response and renders a product page, whereas an \texttt{add} action passes if the verifier can successfully activate an add-to-cart control. A trajectory passes only if every action is reproduced successfully. Each failure report identifies the corresponding trajectory, action, and reason and is supplied to the next code-generation iteration.

\vspace{-0.6em}
\subsection{Generate Synthetic Data}
\label{sec:method:gen}
To reproduce fine-grained multimodal interaction trajectories based on the transformed sessions, we use a LLM-based teacher agent to navigate the simulated environment. This teacher agent needs to plan step-by-step actions while maintaining consistency with the user's high-level intent and persona. It also needs to ensure that each action is valid and realistic before execution, since an incorrect action can irreversibly alter the environment state and cause the trajectory to deviate from the reference behavior. 
We therefore develop a nested reflection--verification architecture: the outer loop maintains task-level coherence across the trajectory, while the inner loop validates each proposed action before execution. See Figure~\ref{fig:main} Section III.

To maintain coherence over multiple interaction steps, the outer loop repeatedly plans, acts, and reflects. The \textbf{plan module} determines the next step using the current web observation and guidance from prior reflections. The \textbf{act module} translates this plan into a browser action with a corresponding rationale. After execution, the \textbf{reflect module} examines the resulting observation and provides high-level guidance for subsequent planning.

To prevent an incorrect action from distorting the remaining trajectory, the \textbf{verify module} in the inner loop evaluates each proposal against three criteria: (1) executability, whether the action can be performed on the current page; (2) trajectory consistency, whether the action aligns with the reference trajectory; and (3) realism, whether a real user would plausibly take the action in the current context. For each criterion, the verifier returns a rationale, score, and confidence value. If an action fails verification, the action module receives the failure reason and generates a revised action.


\section{E-commerce Synthetic Dataset}
\label{sec:method:data}

Using \framework{}, we generate an e-commerce synthetic dataset across five simulated stores covering apparel, food, cookware, toys, and accessories, and four languages: English, French, Turkish, and Hindi.\footnote{The dataset is available at \url{https://huggingface.co/datasets/luyunan/SimTrace}.} Following OPeRA~\citep{wang_opera_2026}, each session data contains three components: action records, web observations, and user context.

\textbf{Action Records.} 
Each session contains an ordered sequence of actions, each with an action type, a natural-language rationale, and action-specific metadata. Table~\ref{tab:action-distribution} shows the action type distribution. Each action type defines its own metadata fields. For example, a \texttt{click} action includes a \texttt{target} field that identifies the selected interface element in the web observation, such as \texttt{"search\_icon"}, whereas a \texttt{type} action includes an \texttt{input} field specifying the text entered by the agent.

\textbf{Web Observations.}
At each step, the dataset records a web observation comprising an HTML representation of the current page and a screenshot. Together, they provide complementary structural and visual context for interpreting the action and its rationale. Following common practice in prior work~\citep{uxagent, sun_llm_2025, wang_opera_2026}, we simplify the raw HTML into a semantic representation that preserves meaningful interface elements.

\textbf{User Context.}
Each session is paired with an inferred persona and intent derived from the raw activity logs following SimGym \citep{li_simgym_2026}. The persona summarizes the user's behavior and preferences, while the intent specifies the session-level goal. See Appendix \ref{app:persona_example} for an example.

Table~\ref{tab:dataset-comparison} compares our dataset with existing public datasets for modeling online user behavior. Large recommendation-oriented datasets~\citep{requena_shopper_2020,ben-shimon_recsys_2015,zhang_large_2015} provide substantial scale but primarily record product-level interactions with limited context. Another line of datasets record real user interactions \cite{lu_can_2026, wang_opera_2026}. They preserve finer-grained user interactions but are typically tied to a specific data-collection platform and do not provide an interactive environment. In contrast, our dataset combines fine-grained interaction context with controllable sandbox environments across multiple platforms, and can be further expanded or customized to new environments. Appendix~\ref{app:data} provides a complete example session.

\begin{table}[t]
    \centering
    \scriptsize

    \begin{minipage}[t]{0.74\columnwidth}
        \vspace{0pt}
        \centering

        \captionof{table}{Comparison of \framework{} with existing datasets. \emph{Multi-platform} indicates whether a dataset spans multiple environment; \emph{Sandbox} indicates whether it is paired with controllable environments for interaction.}
        \label{tab:dataset-comparison}

        \setlength{\tabcolsep}{1.2pt}
        \renewcommand{\arraystretch}{1.08}

        \begin{tabularx}{\linewidth}{
            @{}
            l
            c
            >{\centering\arraybackslash}X
            >{\centering\arraybackslash}X
            c
            c
            c
            @{}
        }
            \toprule
            \textbf{Dataset}
            & \textbf{Size}
            & \textbf{Tasks}
            & \makecell{\textbf{Action}\\\textbf{Space}}
            & \makecell{\textbf{Multi-}\\\textbf{modal}}
            & \makecell{\textbf{Multi-}\\\textbf{platform}}
            & \makecell{\textbf{Sandbox}} \\
            \midrule

            Coveo
            & 203k
            & \makecell{Purchase Prediction}
            & \makecell{6 Semantic Actions}
            & \ding{55}
            & \ding{55}
            & \ding{55} \\

            YOOCHOOSE
            & 9M
            & Recommendation
            & \makecell{Product Identifiers}
            & \ding{55}
            & \ding{55}
            & \ding{55} \\

            Tmall
            & 8M
            & Recommendation
            & \makecell{Product Identifiers}
            & \ding{55}
            & \ding{55}
            & \ding{55} \\

            \midrule

            SHOPCART
            & 31,865
            & \makecell{User Behavior Simulation}
            & \makecell{click, type, terminate}
            & \ding{55}
            & \ding{55}
            & \ding{55} \\

            OPeRA
            & 692
            & All Above
            & \makecell{Rich GUI Actions}
            & \ding{51}
            & \ding{55}
            & \ding{55} \\

            \midrule

            \framework{}
            & 1,513
            & All Above
            & \makecell{Rich GUI Actions}
            & \ding{51}
            & \ding{51}
            & \ding{51} \\

            \bottomrule
        \end{tabularx}
    \end{minipage}
    \hfill
    \begin{minipage}[t]{0.23\columnwidth}
        \vspace{0pt}
        \centering

        \captionof{table}{Action distribution for \framework{}.}
        \label{tab:action-distribution}

        \setlength{\tabcolsep}{1pt}
        \renewcommand{\arraystretch}{1.08}

        \begin{tabular*}{0.82\linewidth}{
            @{\extracolsep{\fill}}
            l
            r
            @{}
        }
            \toprule
            \textbf{Action}
            & \makecell{\textbf{Count}\\\textbf{(\%)}} \\
            \midrule
            Click     & 8,195 (79.40) \\
            Terminate &   901 (8.73)  \\
            Scroll    &   551 (5.34)  \\
            Type      &   295 (2.86)  \\
            Back      &   277 (2.68)  \\
            Select    &    94 (0.91)  \\
            Clear     &     8 (0.08)  \\
            \midrule
            Total     &       10,321   \\
            \bottomrule
        \end{tabular*}
    \end{minipage}
\vspace{-1.2em}
\end{table}

\vspace{-0.4em}
\section{Experiments}
\label{sec:experiments}

We evaluate \framework{} along two complementary perspectives: \textbf{fidelity} and \textbf{downstream task utility}. Fidelity assesses how closely the synthetic data reproduce the characteristics of real data, whereas downstream utility assesses whether they provide effective supervision for user behavior modeling. Specifically, our experiments address the following research questions:
\begin{itemize}
\vspace{-0.5em}
    \item \textbf{RQ1 - Fidelity}: How faithfully does \framework{} reproduce real user behavior compared with existing methods?
    \item \textbf{RQ2 - Downstream Task Utility}: Can synthetic data generated by \framework{} effectively support downstream user behavior modeling tasks?
    \item \textbf{RQ3 - Low-resource Setting}: How effectively can \framework{} support user behavior modeling in low-resource settings, such as next-action prediction? 
\end{itemize}

\subsection{Fidelity (RQ1)}
\label{sec:fidelity}

\vspace{-0.4em}
\paragraph{Experimental setup.}
To evaluate the fidelity of \framework{}, we compare it with prompting and agentic baselines. \emph{Persona} uses the persona representation from SimGym \citep{li_simgym_2026}; \emph{Trajectory} uses a reference trajectory as a one-shot demonstration; and \emph{Trajectory + Persona} combines both. We use UXAgent \citep{uxagent} as the agentic baseline. Its planning--reflection loop conceptually corresponds to the outer loop in \framework{}, providing a practical comparison for the added benefit of action-level verification. For each method, we generate 200 sessions on both \texttt{gemini-3-flash} \citep{google2025gemini3flash} and \texttt{gpt-5.6-sol} \citep{gpt5.6} models.

\vspace{-0.6em}
\paragraph{Evaluation metrics.}
Behavioral fidelity is multifaceted and cannot be adequately characterized by a single metric. We therefore evaluate it at three complementary levels. Outcome-level fidelity uses Jensen–Shannon divergence (JSD) to assess whether synthetic data preserve the distribution of three session outcomes: browsing only, cart abandonment, and checkout. Sequence-level fidelity captures finer-grained trajectory similarity using action-frequency JSD, action transition matrix distance, and trajectory-level Levenshtein distance. In addition to these structural measures, semantic-level fidelity assesses whether generated trajectories reflect user preference similar to that observed in real sessions. Specifically, it evaluates product coherence, diversity, alignment and rationale alignment. 
Together, these metrics capture population-level statistical similarity and session-level behavioral plausibility. Appendix~\ref{app:metrics} provides detailed definitions of each metric.

\vspace{-0.6em}
\paragraph{Analysis.} Table~\ref{tab:fidelity_results_gemini} shows that \framework{} achieves the best fidelity on 7 out of 8 metrics, with consistent gains across all three evaluation levels. We also test with the \texttt{gpt-5.6-sol} model, it largely preserves the method ranking with a slight absolute gain. See Table ~\ref{tab:fidelity_results_gpt} for results.

Beyond the quantitative results, our qualitative analysis reveals two practical issues when generating synthetic trajectories from activity logs. The first issue, which we refer to as \textbf{granularity mismatch}, occurs when activity logs omit the interface-level steps required to reproduce an event. For example, a \texttt{search} event may record only the query, whereas an agent must locate the search field, enter the query, and submit it through multiple actions. The prompt-based \emph{Trajectory} method often advances to the next logged event before completing these intermediate steps, producing discontinuous trajectories. Our outer reflection--planning loop addresses this issue by tracking task progress and expanding each coarse event into the required actions while treating the reference trajectory as high-level guidance. The second issue, which we refer to as \textbf{interface drift}, occurs when actions recorded in an earlier environment are no longer directly executable in the current interface and require intermediate actions to reach the intended state. Without action-level verification, an agent may repeatedly attempt infeasible actions until exhausting its action budget. Our inner verifier instead checks proposed actions before execution, identifies appropriate intermediate steps, and redirects local deviations toward the reference behavior. Together, these mechanisms allow \framework{} to use logged trajectories as flexible behavioral guidance rather than brittle execution scripts.

\begin{table*}[t]
\centering
\caption{
Fidelity comparison between different generation methods on \texttt{gemini-3-flash} model. The best results are shown in \textbf{bold}. Statistically significant improvements ($p < 0.05$) over all other baselines are marked with $^*$.
}
\label{tab:fidelity_results_gemini}
\resizebox{\textwidth}{!}{%
\begin{tabular}{lcccccccc}
\toprule
& \multicolumn{1}{c}{\textbf{Outcome-level}}
& \multicolumn{3}{c}{\textbf{Sequence-level}}
& \multicolumn{4}{c}{\textbf{Semantic-level}} \\
\cmidrule(lr){2-2}
\cmidrule(lr){3-5}
\cmidrule(lr){6-9}

\textbf{Method}
& \shortstack{\textbf{Outcome}\\\textbf{JSD} $\downarrow$\\$[0,1]$}
& \shortstack{\textbf{Action Freq.}\\\textbf{JSD} $\downarrow$\\$[0,1]$}
& \shortstack{\textbf{Transition Matrix}\\$\boldsymbol{L_1}$ \textbf{Distance} $\downarrow$\\$[0,1]$}
& \shortstack{\textbf{Trajectory}\\\textbf{Levenshtein} $\downarrow$\\$[0,1]$}
& \shortstack{\textbf{Product}\\\textbf{Coherence Gap} $\downarrow$\\$[0,1]$}
& \shortstack{\textbf{Product}\\\textbf{Diversity Ratio} $\downarrow$\\$[0,\infty)$}
& \shortstack{\textbf{Product}\\\textbf{Alignment} $\uparrow$\\$[-1,1]$}
& \shortstack{\textbf{Rationale}\\\textbf{Alignment} $\uparrow$\\$[-1,1]$} \\
\midrule

\multicolumn{9}{l}{\textit{Prompting-based methods}} \\

Persona
& $2.94{\times}10^{-1}$
& $5.40{\times}10^{-2}$
& $4.30{\times}10^{-1}$
& $6.74{\times}10^{-1}$
& $3.92{\times}10^{-2}$
& $2.84{\times}10^{-1}$
& $5.75{\times}10^{-1}$
& $6.10{\times}10^{-1}$ \\

Trajectory
& $8.56{\times}10^{-3}$
& $1.10{\times}10^{-2}$
& $2.34{\times}10^{-1}$
& $2.18{\times}10^{-1}$
& $6.86{\times}10^{-2}$
& $\mathbf{7.74{\times}10^{-2}}$
& $5.32{\times}10^{-1}$
& $6.52{\times}10^{-1}$ \\

Trajectory + Persona
& $3.36{\times}10^{-3}$
& $9.73{\times}10^{-3}$
& $2.39{\times}10^{-1}$
& $2.85{\times}10^{-1}$
& $6.85{\times}10^{-2}$
& $9.15{\times}10^{-2}$
& $5.43{\times}10^{-1}$
& $6.54{\times}10^{-1}$ \\

\midrule
\multicolumn{9}{l}{\textit{Agentic methods}} \\

UXAgent
& $1.86{\times}10^{-3}$
& $9.58{\times}10^{-3}$
& $2.30{\times}10^{-1}$
& $3.50{\times}10^{-1}$
& $3.81{\times}10^{-2}$
& $9.68{\times}10^{-2}$
& $5.52{\times}10^{-1}$
& $6.59{\times}10^{-1}$ \\

\textbf{\framework{}}
& $\mathbf{7.98{\times}10^{-5}{}^*}$
& $\mathbf{4.96{\times}10^{-3}}^*$
& $\mathbf{2.27{\times}10^{-1}}$
& $\mathbf{1.80{\times}10^{-1}}$
& $\mathbf{3.75{\times}10^{-2}}$
& $8.50{\times}10^{-2}$
& $\mathbf{5.92{\times}10^{-1}}$
& $\mathbf{6.73{\times}10^{-1}}$ \\

\bottomrule
\end{tabular}
}
\vspace{-1.2em}
\end{table*}

\begin{table}[t]
\centering
\caption{
Downstream performance of models trained on real (TRTR) and synthetic (TSTR) sessions and evaluated on the same held-out real test set. $\Delta$ is TSTR-TRTR. All 95\% CIs for $\Delta$ include zero, indicating no statistically significant differences between two training conditions.
}
\label{tab:downstream_utility}
\fontsize{8.5pt}{8pt}\selectfont
\setlength{\tabcolsep}{5pt}
\begin{tabular}{llcccc}
\toprule
\textbf{Method}
& \textbf{Metric}
& \textbf{TRTR}
& \textbf{TSTR}
& \textbf{$\Delta$}
& \textbf{95\% CI} \\
\midrule

\multicolumn{6}{l}{\textit{Purchase prediction}} \\
\addlinespace[2pt]
SFT (Qwen3.5-4B)
& F1 $\uparrow$
& 64.36
& 65.52
& $+1.16$
& $[-3.00,\,+5.00]$ \\

\midrule
\multicolumn{6}{l}{\textit{Session-based recommendation}} \\
\addlinespace[2pt]

\multirow{4}{*}{RAIN (Graph-based)}
& MRR@5 $\uparrow$
& 12.47
& 12.58
& $+0.11$
& $[-1.59,\,+1.70]$ \\

& HR@5 $\uparrow$
& 21.25
& 20.85
& $-0.40$
& $[-3.20,\,+2.27]$ \\

& MRR@10 $\uparrow$
& 13.20
& 13.64
& $+0.44$
& $[-1.24,\,+2.03]$ \\

& HR@10 $\uparrow$
& 26.71
& 29.12
& $+2.41$
& $[-0.71,\,+5.40]$ \\

\addlinespace[3pt]

\multirow{4}{*}{DIMO (Context-aware)}
& MRR@5 $\uparrow$
& 17.84
& 16.49
& $-1.35$
& $[-3.67,\,+1.16]$ \\

& HR@5 $\uparrow$
& 31.83
& 30.27
& $-1.56$
& $[-5.05,\,+2.28]$ \\

& MRR@10 $\uparrow$
& 18.91
& 18.34
& $-0.57$
& $[-2.85,\,+2.41]$ \\

& HR@10 $\uparrow$
& 39.86
& 40.46
& $+0.60$
& $[-2.83,\,+3.95]$ \\

\bottomrule
\end{tabular}
\vspace{-1.2em}
\end{table}

\vspace{-0.4em}
\subsection{Downstream Task Utility (RQ2)}
\label{sec:downstream_utility}

\vspace{-0.4em}
To assess whether the synthetic sessions generated by our pipeline provide useful training signals across multiple downstream tasks. We evaluate \framework{} on two tasks: purchase prediction and session-based recommendation.
For each task, we compare models trained and tested on real sessions (TRTR) with models trained on synthetic sessions and tested on real sessions (TSTR). Both conditions use the same held-out real test set, isolating the effect of replacing real training trajectories with their synthetic counterparts.

\vspace{-0.6em}
\paragraph{Purchase Prediction}
Given the interaction history before a session's terminal event, the model predicts whether the session ends in a purchase. We evaluate this binary classification task using the F1 score. For training, we use all available matched real--synthetic session pairs and balance the purchase and non-purchase classes, yielding 739 examples in each training condition. To construct the held-out test set, following prior work~\citep{requena_shopper_2020}, we exclude real sessions with length $L<5$ to ensure sufficient interaction history. We then randomly sample 500 of the remaining sessions to form the final held-out test set.

We finetune \texttt{Qwen3.5-4B} \citep{qwen3.5} with QLoRA for five epochs with a learning rate of $1 \times 10^{-5}$. We report preliminary results averaged over three random seeds, together with the TSTR--TRTR difference and its 95\% confidence interval.

\vspace{-0.6em}
\paragraph{Session-Based Recommendation}
Let $\mathcal{V}$ denote the item set, and let $S=(v_1,v_2,\ldots,v_L)$ denote a sequence of item interactions within a user session. Given an observed session $S$, session-based recommendation aims to predict the next item $v_{L+1}\in\mathcal{V}$ by ranking all candidate items and returning the top-$K$ items. Following prior work~\citep{zhang_beyond_2023}, we remove sessions containing only one interaction and items appearing fewer than five times. Applying these filters to the matched real and synthetic data yields 357 training sessions covering 110 items. Since ID-based recommendation models cannot score items unseen during training, we restrict the held-out real test set to the shared training-item vocabulary. The resulting test set contains 434 sessions involving 40 items.

We apply it to two complementary recommendation methods. RAIN \citep{zeng2025rain} is a recent graph-based session recommendation method with publicly available code. DIMO \citep{zhang_disentangling_2024} is a context-aware method that additionally incorporates product titles, vendors, product types, and images into its recommendation model. DIMO therefore allows us to assess whether the rich multimodal content generated by our pipeline provides additional utility for downstream recommendation. We evaluate both models using Hit Rate at $K$ (HR@$K$) and Mean Reciprocal Rank at $K$ (MRR@$K$), where $K\in\{5,10\}$. We repeat each experiment with three random seeds and use bootstrap resampling to estimate the TSTR--TRTR differences and 95\% confidence intervals.

\vspace{-0.6em}
\paragraph{Findings}
Table \ref{tab:downstream_utility} reports downstream performance for the models trained on real and synthetic sessions using the same held-out real test set. It shows that every 95\% confidence interval includes zero, indicating no statistically significant difference between the two training conditions under our experimental setting. These results suggest that the synthetic sessions preserve task-relevant training signals for both purchase prediction and session-based recommendation.

\begin{figure}[h]
\begin{center}
\includegraphics[scale=0.40]{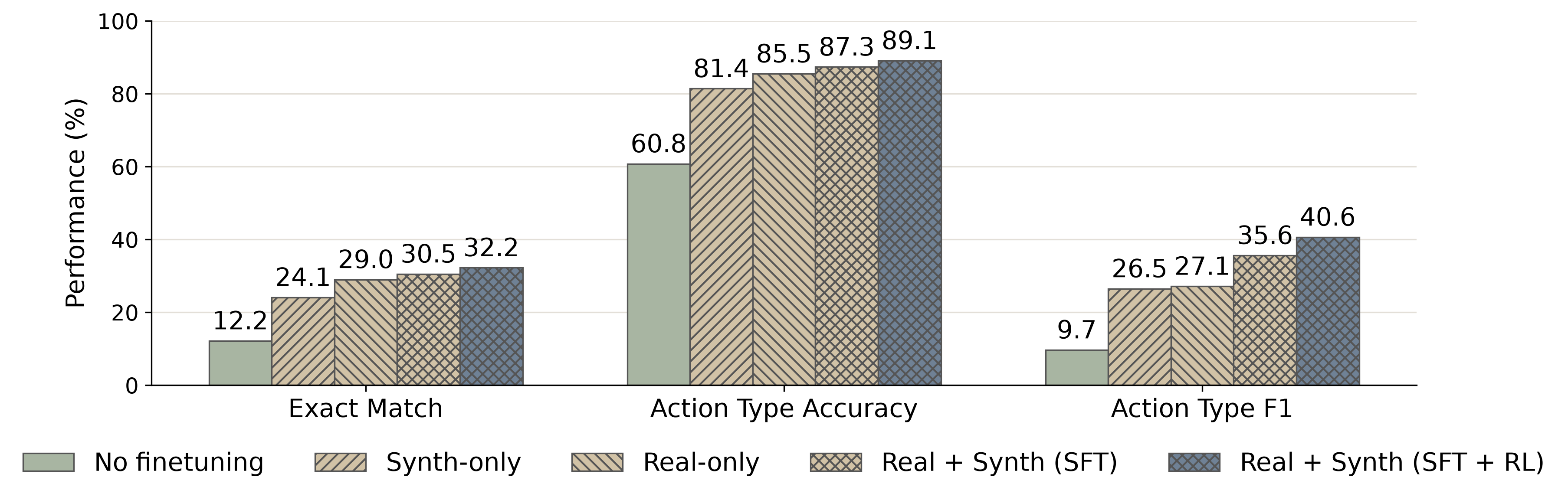}
\end{center}
\vspace{-0.4em}
\caption{Next action prediction performance of \texttt{Qwen3.5-9B} on the OPeRA test set. Augmenting real with synthetic data improves all metrics, with RL achieving the best overall performance.}
\label{fig:user_modeling}
\vspace{-1.2em}
\end{figure}

\vspace{-0.4em}
\subsection{Low-Resource Setting (RQ3)}
\label{sec:low_resource}
\vspace{-0.4em}


For small businesses, limited traffic hinders the training of user simulation models that can perform fine-grained navigation behaviors and enable practitioners to analyze how users interact with a website step by step. Therefore, we assess how effectively synthetic data generated by \framework{} supports next-action prediction in low-resource settings. Specifically, we sample 50 sessions from OPeRA~\citep{wang_opera_2026} as the real training data and compare real-only training (Real-only) with synthetic-only (Synth-only) and synthetic augmentation (Real + Synth) training. All these trainings use \texttt{Qwen3.5-9B} \citep{qwen3.5} as the base model and are evaluated on the held-out OPeRA test set using exact match, action-type F1, and action-type accuracy. A prediction counts as an exact match only when both its \texttt{action} and \texttt{target} fields exactly match the reference. 




\vspace{-0.6em}
\paragraph{SFT Training and Analysis.}

We formulate each session as a multi-turn conversation, where each user turn contains the current web observation and each assistant turn contains a rationale followed by a structured action. The system prompt specifies the task instruction, persona, and shopping intent. See Appendix~\ref{app:sft_prompt} for the complete template and Appendix \ref{app:training_detail} for training details. 

Figure~\ref{fig:user_modeling} shows that SFT on synthetic data improves the base model without finetuning by 11.9--20.6 percentage points across the three metrics. Training on the 50 real sessions yields stronger performance, likely because of distributional differences between synthetic and real supervision, as analyzed in Appendix~\ref{app:user_modeling:SR}. Despite this gap, synthetic data provides complementary supervision: jointly training on real and synthetic sessions produces further gains of 1.5--8.5 percentage points over real-only. 
The improvement mainly comes from broader \texttt{target} field coverage, which reduces overprediction of frequent values and helps the model recognize infrequent or previously unseen interface elements. See Appendix \ref{app:user_modeling:data} for details.

However, \texttt{target} prediction remains the main bottleneck for exact-match performance.
Target field values are long and highly sparse: 77.4\% of unique targets occur only once in OPeRA, and 14.9\% contain at least 200 characters. Nevertheless, their hierarchical structure encodes behavioral similarity. For example, \texttt{"customers\_also\_bought.<product1>"} and \texttt{"customers\_also\_bought.<product2>"} both indicate exploration of recommended alternatives, whereas a target beginning with \texttt{"review"} indicates further investigation of the current product. Exact match treats these predictions as equally different, despite the first two reflecting similar behavioral directions. We therefore introduce an RL objective that assigns partial credit according to hierarchical similarity of \texttt{target} field.

\vspace{-0.6em}
\paragraph{RL Training and Analysis.}
Following the approach of \citep{deepseek-ai_deepseek-r1_2025}, we initialize the training with SFT on synthetic data to learn the dependencies among web context, action type, and target. We then apply RL on the real sessions to align the learned policy with the behavioral patterns and target distribution of the real environment. Each real session is converted into $\langle\text{context},\text{action}\rangle$ pairs, where the context contains all preceding web observations and executed actions and the prediction is the structured next action.

Based on our observations from the SFT results, we design a reward function for the \texttt{target} field that assigns partial credit based on the prediction. The reward comprises three terms: a string-similarity reward, a binary validity reward indicating whether the predicted target exists in the current UI context, and an exact-match bonus.
Let $\hat t$ and $t$ denote the predicted and reference targets: 
\[
R_{\mathrm{target}}
=
w_{sub} {s_\mathrm{sub}}(\hat t,t)
+
w_{valid} \mathbb{I}[\hat t\in\mathcal{T}(x_t)]
+
w_{exact} \mathbb{I}[\hat t=t],
\]
where $s_{\mathrm{sub}}(\hat{t},t)$ is the symmetric longest-common-substring ratio between predicted and ground-truth target, and $\mathcal{T}(x_t)$ is the set of semantic identifiers in the current web observation.

Target accuracy alone, however, does not ensure the prediction accuracy. Following \citet{zhang_shop-r1_2026}, the full reward therefore combines $R_{\mathrm{target}}$ with $R_{\mathrm{action}}$, which rewards correct action-type predictions, and $R_{\mathrm{format}}$, which rewards valid structured responses. We optimize this multi-reward objective using Group Reward-Decoupled Normalization Policy Optimization~\citep{liu_gdpo_2026}; implementation details are provided in Appendix~\ref{app:training_detail}. 
Relative to mixed-data SFT, the RL stage further improves the three metrics by 1.7\%, 1.8\%, and 5.0\%, respectively. 
The improvements mainly come from better output validity and target selection. RL eliminates malformed JSON predictions and more often redirects predictions from an unrelated UI component to the correct hierarchical target path. For example, it corrects \texttt{reviews.popover.review\_images.next} to the ground-truth \texttt{buybox.purchase\_form.add\_to\_cart}. It also improves predictions when the correct parent path is identified, but the final target segment is wrong. See Appendix~\ref{app:user_modeling:rl} for details.

\section{Related Work}
\label{sec:related}

\vspace{-0.6em}
\paragraph{LLM-Based User Simulation}

Recent LLM-based user simulation can be categorized into two paradigms: prompting LLMs and training dedicated user models. Among prompt-based methods, a large amount of work focuses on steering LLM behavior through personas. PAARS \citep{mansour_paars_2025} induces shopper personas from anonymized sessions, while SimGym \citep{li_simgym_2026} constructs buyer profiles for offline A/B testing. Other work enriches simulators with domain knowledge. SAGE grounds conversations in business principles \citep{shea_sage_2025}, whereas VISTA enables interaction through both UI and API actions \citep{lu_vista_2026}. Although flexible, prompting-based methods rely on powerful LLMs, making long multimodal simulations costly. 
Dedicated user models reduce this cost by learning human behavior into model parameters. UserLM \citep{naous_flipping_2025} and Socratic \citep{kong_platolm_2024} learn from human-authored turns to reproduce realistic questions, while Shop-R1 \citep{zhang_shop-r1_2026} and Customer-R1 \citep{wang_customer-r1_2025} fine-tune models to create virtual shoppers for e-commerce domain. However, these models remain tied to their training domains and interaction protocols. In contrast, \framework{} generates environment-specific supervision for training efficient user models tailored to the given web environment.

\vspace{-0.6em}
\paragraph{Datasets for User Behavior Modeling}

User behavior modeling requires data that captures both user actions and the context in which they occur. Existing datasets generally fall into two categories. Large-scale datasets contain millions of shopping or recommendation sessions but typically represent them as sequences of item identifiers or coarse event types, omitting the interface observations and semantic context needed to reconstruct user decisions \citep{requena_shopper_2020,ben-shimon_recsys_2015,zhang_large_2015}. Context-rich datasets preserve semantic actions, webpage content, but are costly to collect, limited in scale, and usually tied to a single platform \citep{wang_opera_2026,sun_llm_2025}. \framework{} bridges this gap by transforming private logs into standardized, anonymized trajectories and generating scalable multimodal clickstreams of the given environment.


\section{Conclusion}
\label{sec:conclusion}

We introduce \framework{}, a scalable framework for generating semantically faithful multimodal computer-use client-agent trajectories for user modeling. \framework{} transforms real activity logs into standardized and anonymized trajectories, constructs a simulated twin environment grounded on the given web, and uses a teacher agent to generate behaviorally grounded interactions. 
Our evaluation shows that \framework{} reproduces real user behavior more faithfully than the baselines. Besides, models trained on synthetic data perform comparably to those trained on real data for multiple downstream tasks. Augmenting real data with synthetic data further improves next-action prediction accuracy by 11.0\%. These results demonstrate the potential of \framework{} to alleviate real-data scarcity and reduce reliance on sensitive user logs when training downstream models.
\section{Limitations}
\label{sec:limitations}

The proposed data generation method is designed to be domain-adaptive, while our empirical evaluation focuses on an e-commerce scenario. Further validation is needed in other domains, where user semantic action taxonomy, privacy requirements may differ. Future work should evaluate \framework{} in domains such as finance, education, and healthcare.
Secondly, \framework{} does not fully solve the cold-start problem, as trajectory-conditioned generation still requires initial activity logs. Although synthetic data generated from a similar environment can provide useful supervision when target-domain data are scarce, our results show that such data remain less effective than real target-domain interactions for next-action prediction. Future work should investigate cross-environment schema alignment and minimal-data adaptation to bootstrap generation from related environments while reducing reliance on target-domain logs.

\subsubsection*{Acknowledgments}
This research was supported by grant funding from Shopify and Toloka. We gratefully acknowledge both organizations for their generous support and thank our collaborators for their constructive feedback and valuable discussions throughout the project, which helped shape the research direction and strengthen this work.

\bibliography{iclr2027_conference}
\bibliographystyle{iclr2027_conference}

\clearpage
\appendix

\section{Representative Session Selection}
\label{app:session-selection}
Let \(S\) denote the candidate sessions. Each session \(s\) is associated with an outcome \(y_s\), an interaction-sequence length \(l_s\), and a set of encountered products \(p_s\). 
Our objective is to select a subset of sessions that maximizes product coverage subject to two constraints: (1) the selected sessions are evenly distributed across the outcomes, and (2) the \(l_s\) are larger than 3 for meaningful interactions and smaller than the empirical 95th-percentile length constraint to avoid unusually long sessions.

We solve this selection problem using a greedy algorithm. We first filter the candidate sessions according to the sequence-length constraint (e.g.  \(3 \leq l_s \leq 20\)). At each iteration, we consider each outcome and select the eligible session with the largest marginal coverage of previously unseen products.
The procedure terminates when it reaches the specified sample-size budget or a configured product-coverage threshold, such as 95\%. See Algorithm~\ref{alg:session-selection} for pseudocode.

\begin{algorithm}[h]
\caption{Session Selection}
\label{alg:session-selection}
\begin{algorithmic}[1]
\Require Candidate sessions \(S\); optional budget \(B\); product coverage threshold \(\tau\); minimum length \(l_{\min}\); maximum length \(l_{\max}\)
\Ensure Selected sessions \(X\)

\State \(S' \gets \{s\in S:l_{\min}\leq l_s\leq l_{\max}\}\)
\State \(U\gets\bigcup_{s\in S'}P_s\) \Comment{\(U\) is product union}
\State \(\mathcal{Y}\gets
\{\text{Browser},\text{Cart Abandoner},\text{Checkout}\}\)
\State \(X\gets\emptyset,\ C\gets\emptyset\)

\While{\((B\neq\emptyset \textbf{ and } |X|<B)
       \textbf{ or } (B=\emptyset \textbf{ and } |C|/|U|<\tau)\)} \Comment{\(C\) is the covered product list}
    \For{\(y\in\mathcal{Y}\)}
        \State \(A_y\gets
        \{s\in S'\setminus X:y_s=y\}\)
        \Comment{\(A_y\) is available sessions for outcome \(y\)}

        \If{\(A_y=\emptyset\)}
            \State \Return \(X\) with an infeasibility warning
        \EndIf

        \State \(s_y^*\gets
        \operatorname*{arg\,max}_{s\in A_y}|P_s\setminus C|\)
        \State \(X\gets X\cup\{s_y^*\}\)
        \State \(C\gets C\cup P_{s_y^*}\)
    \EndFor
\EndWhile

\State \Return \(X\)
\end{algorithmic}
\end{algorithm}

\begin{table*}[h]
\centering
\caption{
Fidelity comparison between different generation methods on the
\texttt{gpt-5.6-sol} model. The best results are shown in
\textbf{bold}. Statistically significant improvements ($p < 0.05$) over all other baselines are marked with $^*$.
}
\label{tab:fidelity_results_gpt}
\resizebox{\textwidth}{!}{%
\begin{tabular}{lcccccccc}
\toprule
& \multicolumn{1}{c}{\textbf{Outcome-level}}
& \multicolumn{3}{c}{\textbf{Sequence-level}}
& \multicolumn{4}{c}{\textbf{Semantic-level}} \\
\cmidrule(lr){2-2}
\cmidrule(lr){3-5}
\cmidrule(lr){6-9}

\textbf{Method}
& \shortstack{\textbf{Outcome}\\\textbf{JSD} $\downarrow$\\$[0,1]$}
& \shortstack{\textbf{Action Freq.}\\\textbf{JSD} $\downarrow$\\$[0,1]$}
& \shortstack{\textbf{Transition Matrix}\\
  $\boldsymbol{L_1}$ \textbf{Distance} $\downarrow$\\$[0,1]$}
& \shortstack{\textbf{Trajectory}\\
  \textbf{Levenshtein} $\downarrow$\\$[0,1]$}
& \shortstack{\textbf{Product}\\
  \textbf{Coherence Gap} $\downarrow$\\$[0,1]$}
& \shortstack{\textbf{Product}\\
  \textbf{Diversity Ratio} $\downarrow$\\$[0,\infty)$}
& \shortstack{\textbf{Product}\\
  \textbf{Alignment} $\uparrow$\\$[-1,1]$}
& \shortstack{\textbf{Rationale}\\
  \textbf{Alignment} $\uparrow$\\$[-1,1]$} \\
\midrule

\multicolumn{9}{l}{\textit{Prompting-based methods}} \\

Persona
& $1.94{\times}10^{-1}$
& $4.39{\times}10^{-2}$
& $3.82{\times}10^{-1}$
& $6.08{\times}10^{-1}$
& $6.06{\times}10^{-2}$
& $1.03{\times}10^{-1}$
& $5.97{\times}10^{-1}$
& $\mathbf{6.47{\times}10^{-1}}$ \\

Trajectory
& $3.98{\times}10^{-3}$
& $1.65{\times}10^{-2}$
& $2.28{\times}10^{-1}$
& $2.59{\times}10^{-1}$
& $7.80{\times}10^{-2}$
& $6.45{\times}10^{-2}$
& $5.41{\times}10^{-1}$
& $6.38{\times}10^{-1}$ \\

Trajectory + Persona
& $1.20{\times}10^{-3}$
& $2.03{\times}10^{-2}$
& $2.38{\times}10^{-1}$
& $2.80{\times}10^{-1}$
& $8.00{\times}10^{-2}$
& $7.14{\times}10^{-2}$
& $5.89{\times}10^{-1}$
& $6.34{\times}10^{-1}$ \\

\midrule
\multicolumn{9}{l}{\textit{Agentic methods}} \\

UXAgent
& $7.69{\times}10^{-4}$
& $1.25{\times}10^{-2}$
& $2.53{\times}10^{-1}$
& $3.71{\times}10^{-1}$
& $1.00{\times}10^{-1}$
& $\mathbf{2.58{\times}10^{-2}}$
& $5.92{\times}10^{-1}$
& $6.07{\times}10^{-1}$ \\

\textbf{\framework{}}
& $\mathbf{0.00^*}$
& $\mathbf{5.59{\times}10^{-3}}$
& $\mathbf{2.27{\times}10^{-1}}$
& $\mathbf{2.39{\times}10^{-1}}$
& $\mathbf{5.79{\times}10^{-2}}$
& $3.90{\times}10^{-2}$
& $\mathbf{6.18{\times}10^{-1}}$
& $6.41{\times}10^{-1}$ \\

\bottomrule
\end{tabular}%
}
\end{table*}

\section{Simulate Twin Environment}
\label{app:env-gen}

We construct a simulated twin of each original website using an autonomous exploration and generation pipeline adapted from ShopGym~\citep{savadikar_shopgym_2026}. The objective is to reproduce the visual organization and interaction patterns that shape the user's navigation experience, while allowing identifying content to be replaced with synthetic alternatives due to sensitivity information. Exploration produces specifications supported by browser evidence. Then the generation module uses these specifications and the captured or synthetic static data to implement a runnable sandbox. Below, we describe the exploration components, the specification representation, and the generation procedure.

\subsection{Exploration}
\label{app:env-gen-exploration}

\paragraph{Domain-specific interface-area Configuration.}
To reconstruct a website faithfully, the agent must first identify both its visual structure and its interactive behavior. We therefore formulate exploration as a configurable inspection process that specifies which interface areas the agent should examine and what evidence it should collect. This configuration serves as a coverage checklist for the coding agent. While some interface areas are common across domains, such as the homepage, header, and navigation, others depend on the target website. The exploration configuration can therefore be adapted to different website domains.

In our e-commerce setting, we configure the following nine interface areas:
\begin{enumerate}
    \item \textbf{Homepage:} the hero region and subsequent sections, including their ordering, layouts, content, and interactive elements.
    \item \textbf{Header and navigation:} header structure, announcement bars.
    \item \textbf{Cart:} its presentation as a drawer, popup, or page, together with empty and populated states, quantity changes, and item removal.
    \item \textbf{Product pages:} image galleries, product information, variant selectors, quantity controls, availability signals, and recommendations.
    \item \textbf{Collections:} product-card layouts, filtering, sorting, and pagination.
    \item \textbf{Search:} the search entry point, predictive suggestions, and the presentation of results.
    \item \textbf{Footer:} link groups, policy links, payment indicators, and social links.
    \item \textbf{Information pages:} shipping, returns, privacy, terms, frequently asked questions, and contact and about pages when present.
    \item \textbf{Floating interfaces:} cookie banners, newsletter popups, chat widgets, and age gates, including their appearance and dismissal behavior.
\end{enumerate}
The exploration plan includes all configured areas. When a feature is absent or an interaction cannot be exercised, the agent explicitly records this outcome rather than inferring its behavior. The agent may also append new exploration tasks when it encounters functionality not covered by the predefined configuration.

\paragraph{Interactive inspection.}
The coding agent executes the exploration plan using Playwright browser automation. Rather than inspecting only static page layouts, it also exercises relevant interactions to capture state transitions. For example, cart inspection includes adding an item, changing its quantity, and removing it, while search inspection includes entering query prefixes and observing the suggestion panel.
For each distinct interface state, the agent records a full-page screenshot together with the corresponding accessibility snapshot. We organize this evidence by exploration task so that each observation remains associated with the interface area, interaction, and state that produced it.

\paragraph{Structured properties recording.}
Screenshots alone do not fully specify how an interface behaves. We therefore complement visual evidence with structured descriptions of each observed element. The agent records six properties: \emph{type}, \emph{layout}, \emph{content}, \emph{behavior}, \emph{styling}, and \emph{UX notes}. These fields capture the element's functional role, spatial arrangement, displayed information, response to user interaction, visual appearance, and relevant edge cases. For example, the description of a navigation panel records its opening trigger, screen position, entry hierarchy, and dismissal behavior. This representation makes interactive functionality and state transitions explicit while retaining screenshots as visual references.

\subsection{Specifications}
\label{app:env-gen-specification}

The exploration outputs are consolidated into a specification manual with three complementary components. First, a natural-language design manual summarizes the overall website content, structure and visual style. Second, a structured element-level specification records interface-specific properties for each area defined in the exploration configuration, including layout, navigation-menu design, filtering dimensions, pagination behavior, product-option controls, etc. Third, we retain prefetched static data files, such as product and collection JSON files, which provide the structured content required to populate the reconstructed interface.

Together, these components summarize the website's interface structure and behavior from its underlying content and provide the coding agent with sufficient information to reproduce the observed environment.

\subsection{Generation}
\label{app:env-gen-generation}
Given the resulting specification, the generation stage constructs a simulated twin that preserves the structure and interaction patterns of the original environment while replacing sensitive vendor-specific content.

\paragraph{Synthetic content generation.}
In our e-commerce setting, we treat vendor-specific product information as sensitive content, including product titles, descriptions, vendor identities, and images. We therefore transform the prefetched static product data before using it in the reconstructed website.

For each product, we first identify its original vendor information and generate a synthetic vendor identity. We then regenerate textual product content, including titles and descriptions, conditioned on the synthetic vendor identity while preserving the product's underlying semantic attributes.
Product images require additional care because directly transforming the original images may retain identifying or copyrighted visual content. We therefore use a two-stage generation pipeline. The first model produces a textual description of the original product image. A second image-generation model then receives only this description and the synthetic vendor information and generates a new image without access to the original image. This separation preserves high-level product semantics while reducing direct visual reproduction of the source content.

\paragraph{Stepwise source-code generation.}
Generating the entire website in a single step makes it difficult to maintain consistency and localize errors. Instead, the coding agent implements the configured interface areas incrementally. For each generation task, it receives the relevant portion of the design manual, the associated structured specification and static data, and the current source code. It then implements the corresponding layouts, content, and interactions.

The generation plan, source code, and intermediate artifacts persist across iterations. Consequently, later tasks can build directly on previously implemented components, while errors can be corrected locally without regenerating the complete website. The final application, together with its synthetic content and supporting data, constitutes the simulated twin used for downstream user simulation.

\subsection{Cost}
\label{app:env-gen-cost}

Constructing a simulated twin requires approximately 28--30M tokens and 2.9--4.2 hours end-to-end using the Pi harness with GPT-5.5, with an estimated API cost of \$60--64 per website. Token usage includes cached input, uncached input, and output tokens.

\section{Dataset}

\subsection{Persona Example}
\label{app:persona_example}

\begin{tcolorbox}[
  enhanced,
  colback=gray!5!white,
  colframe=gray!60!black,
  title=Persona Example,
]
\small
\begin{verbatim}
{
    "behavioral": {
        "exploration_depth": 0.45,
        "price_sensitivity": "mid-range"
    },
    "values": {
        "ethics": 0.1,
        "performance": 0.5,
        "premium": 0.2
    },
    "reasoning": {
        "ethics": "No indicators of environmental or ethical
        considerations (e.g., non-toxic plastics, sustainable
        packaging, or local manufacturing) were found in the
        product data. Score: 0.1",

        "exploration_depth": "The user spent a significant amount
        of time (avg 347.5s) relative to the number of products
        viewed (2), suggesting they were carefully reading
        descriptions rather than browsing broadly. Score: 0.45",

        "performance": "100% of browsed products focus on the tactile
        performance and functional outcomes ('slow-rise experience', 
        'groovy tactile experience', 'stress relief'). Focus is on 
        how the product functions as a sensory tool. Score: 0.5
        (max for browsing only)",

        "premium": "Description mentions 'velvety texture' and 
        'perfected the formula', signaling a preference for 
        established brand quality over generic toys, though it 
        stops short of luxury or exclusive keywords. Score: 0.2",

        "price_sensitivity": "Browsed price bucket is standard for 
        name-brand sensory fidget toys. There is no evidence of 
        extreme budget-seeking or high-end collector-grade pricing.
        Classification: 'mid-range'"
    },
    "confidence": {
        "behavioral": 0.4,
        "values": 0.2
    }
}
\end{verbatim}
\end{tcolorbox}

\subsection{Dataset Details}
\label{app:data}

We use \texttt{gemini-3-flash}~\citep{google2025gemini3flash} to generate the e-commerce synthetic dataset. We release the dataset under the Creative Commons Attribution-NonCommercial 4.0 International license (CC BY-NC 4.0). The following example illustrates the structure of a complete session in the generated dataset. For readability, the \texttt{simplified\_dom} and \texttt{persona} fields are truncated.

\begin{lstlisting}[style=json]
{
  "store_id": "68e1563a-676a-694b-ce31-820f79080bad",
  "session_id": "d79f58cc-30fc-046f-bc52-38014c1642a7",
  "user_id": "4c5381ad-ab85-6e0d-ca9f-71ce393443f1",
  "timestamp": "2026-07-01T21:56:48.861744",
  "action_type": "click",
  "target": "needoh",
  "rationale": "I'll start by checking out the NeeDoh brand collection, as I'm looking for 
                high-quality sensory toys for my collection.",
  "input_text": null,
  "simplified_dom": "<html><body parser-is-focused=\"true\">
                     ...
                     <a href=\"/collections/novelty-toys\"
                        parser-semantic-id=\"needoh\"
                        parser-clickable=\"true\">
                       <span>NeeDoh</span>
                     </a>
                     ...
                     </body></html>",
  "action_json": {
        "action": "click",
        "description": "Clicking on the NeeDoh brand collection.",
        "rationale": "I'll start by checking out the NeeDoh brand
                      collection, as I'm looking for high-quality sensory
                      toys for my collection.",
        "target": "needoh",
        "url": "/collections/novelty-toys"
    },
  "image": "41fd9aa7f6e8c0e5f260c3da.png",
  "intent": "I'm looking for Sensory Toys, greeting cards, sensory toys, Stuffed Animals."
  "persona": {
        "behavioral": {
            "exploration_depth": 0.45,
            "price_sensitivity": "mid-range"
            }, 
        "values": {
            "ethics": 0.1,
            "performance": 0.5,
            "premium": 0.2
            }, 
        "reasoning": {...}, 
        "confidence": {...}
    }
}
\end{lstlisting}

\section{Fidelity Metrics}
\label{app:metrics}

This section defines the eight fidelity metrics reported in Table~\ref{tab:fidelity_results_gemini} and Table~\ref{tab:fidelity_results_gpt}. 
Let $A$ denote the semantic action set defined in Table~\ref{tab:action_taxonomy}, with $|A|=9$. 
This standardized taxonomy abstracts away website-specific actions and enables the calculation of distributional metrics, such as Jensen--Shannon divergence (JSD) and Transition Matrix L1 Distance.

\textbf{Outcome JSD} measures the divergence between the session outcome distributions of real and synthetic data. We categorize each session as \emph{Browser}, \emph{Cart Abandoner}, or \emph{Checkout}. Jensen--Shannon divergence (JSD) \citep{lin_divergence_1991} is widely used to evaluate user simulators by quantifying the discrepancy between simulated and real user behavior \citep{xia_advancing_2024}. Using base-2 logarithms, the score ranges from 0 to 1, with lower values indicating more similar outcome distributions.

\textbf{Action Frequency JSD} measures the divergence between the marginal action distributions of real and synthetic data. We pool all actions in each corpus into a normalized frequency vector over $A$ and compute the JSD between the two vectors. Unlike Outcome JSD, which considers only terminal outcomes, this metric captures how frequently each action occurs throughout a session. For example, a generator may reproduce the real checkout rate while omitting the \texttt{search} and \texttt{detail} actions that typically precede a purchase. The score ranges from 0 to 1, with lower values indicating more similar action distributions.

\textbf{Transition Matrix $L_1$ Distance} measures the similarity of local action-to-action dynamics following prior work \citep{lu_vista_2026}. For each corpus, we estimate a row-normalized first-order transition matrix $P\in[0,1]^{|A|\times|A|}$, where $P_{ij}=\Pr(a_{t+1}=j\mid a_t=i)$. We compute the normalized $L_1$ distance as
$\frac{1}{|A|}\sum_{i}\frac{1}{2}\sum_{j}\lvert P^{\mathrm{real}}_{ij}-P^{\mathrm{syn}}_{ij}\rvert$. This metric captures sequential structure that marginal action frequencies cannot distinguish. For example, replacing every \texttt{detail}$\rightarrow$\texttt{add} transition with \texttt{add}$\rightarrow$\texttt{detail} preserves action frequencies but increases the transition distance. The score ranges from 0 to 1, with lower values indicating more similar transition patterns.

\textbf{Trajectory Levenshtein} measures the normalized Levenshtein edit distance between the semantic action sequences of real and synthetic sessions, following prior work \citep{sun_llm_2025}. A score of 0 indicates that the agent replayed its reference trajectory exactly.

\textbf{Product Coherence Gap} measures whether synthetic sessions maintain the same degree of topical consistency as real sessions. For each session containing at least two products, we compute the mean pairwise cosine similarity between the embeddings of the viewed products. We rescale the value to $[0,1]$. We then report the absolute difference between the mean coherence scores of the real and synthetic corpora. The lower values indicating more similar within-session coherence. A large gap in either direction suggests that the agent either browses across unrelated product categories or focuses more narrowly than real users.

\textbf{Product Diversity Ratio} measures the relative difference in overall product coverage between the real and synthetic corpora. Let $n^{\mathrm{real}}$ and $n^{\mathrm{syn}}$ denote the numbers of unique products browsed across all real and synthetic sessions, respectively. We compute $\lvert n^{\mathrm{syn}}/n^{\mathrm{real}}-1\rvert$. A score of 0 indicates that the synthetic corpus covers the same number of unique products as the real corpus, with lower values indicating more similar product diversity. Although this metric is unbounded in principle, the synthetic corpora in our experiments consistently cover fewer products than the real data. This result highlights the need to increase product diversity in future synthetic data generation.

\textbf{Product Alignment} measures whether the real and synthetic sessions involve semantically similar products. For each product in a synthetic session, we retrieve the most similar product from the matched real session using inner-product search over $L_2$-normalized embeddings and average the resulting values within each session across all matched pairs. The score ranges from -1 to 1, with higher values indicating stronger product alignment.

\textbf{Rationale Alignment} measures whether matched real and synthetic sessions reflect similar underlying shopping intents. Given the action sequence and product context of each session, an LLM generates a natural-language description of the inferred intent. We embed these descriptions and report the mean cosine similarity between the descriptions of each matched real--synthetic pair. This metric captures semantic agreement despite differences in specific actions or products; for example, two sessions may follow different trajectories while both reflecting comparison shopping for a budget-friendly gift.


\section{User Model System Prompt}
\label{app:sft_prompt}

The template below is the system message used for training and evaluation. The \texttt{\{persona\}} and \texttt{\{intent\}} fields are filled per session. The user turn supplies the current observation as \texttt{\# context} followed by the HTML of the page, and assistant turn is the single JSON object the template requires.

\begin{tcolorbox}[
  enhanced,
  breakable,
  colback=gray!5!white,
  colframe=gray!60!black,
  title=User Model System Prompt Template,
]
\small
\begin{verbatim}
You pretend to be a user browsing the store website and do
shopping based on your intent. Your task is to predict the next
action and provide rationale for the action based on your
persona, intent, previous actions and context. The history
action (with details described below), rationale, context and
the user persona will be provided to you.

# Action Space
Each action object must include an `action` key specifying one
of the following types. Include required fields exactly as
shown.
## Click
{ "action": "click", "target": "<element_semantic_id>",
  "description": "Clicking ..." }
## Type (with optional Enter submit)
{ "action": "type", "target": "<input_semantic_id>",
  "text": "<text>", "enter": true,
  "description": "Typing and submitting ..." }
## Select (e.g., dropdowns)
{ "action": "select", "target": "<select_semantic_id>",
  "value": "<option_value>", "description": "Selecting ..." }
## Clear (clear an input field)
{ "action": "clear", "target": "<input_semantic_id>",
  "description": "Clearing ..." }
## Scroll (scroll the chat window or current page)
{ "action": "scroll",
  "target": "<optional_element_semantic_id>",
  "direction": "up", "amount": 300,
  "description": "Scrolling up ..." }
{ "action": "scroll",
  "target": "<optional_element_semantic_id>",
  "direction": "down", "amount": 300,
  "description": "Scrolling down ..." }
## Navigation (use when the chatbot provides a URL to open)
{ "action": "goto_url", "url": "https://example.com",
  "description": "Navigating ..." }
{ "action": "back",    "description": "Going back ..." }
{ "action": "forward", "description": "Going forward ..." }
{ "action": "refresh", "description": "Refreshing ..." }
## Terminate (only if explicitly instructed by the step)
{ "action": "terminate", "description": "Terminating ..." }

# Rationale
The rationale is a first-person sentence (<=25 words)
explaining why you are taking the action. Do not mention HTML,
tag ids, or the simulation.

# Context
Your context will be a HTML of the webpage you are looking at.

# Persona
The user persona reflects the user's price sensitivity,
exploration and preference.
Here is your persona:
{persona}

# Intent
Here is your intent:
{intent}

# Output Format
You need to predict the next action and provide rationale for
the action. Your output should be a single, flat JSON object
with `rationale`, `action`, and the type-specific fields shown
above. For example:
{"rationale": "I want to search for a necklace.",
 "action": "type", "target": "search", "text": "necklace",
 "enter": true, "description": "Typing and submitting necklace
 in the search bar."}

<IMPORTANT>
OUTPUT A SINGLE JSON OBJECT, NOTHING ELSE.
</IMPORTANT>
\end{verbatim}
\end{tcolorbox}

\section{Discussion for Next-Action Prediction}
\label{app:user_modeling}

\subsection{Training Details}
\label{app:training_detail}

We use \texttt{Qwen3.5-9B} as the base model for all next-action prediction experiments. Both SFT and RL use full-parameter fine-tuning with BF16 mixed precision and gradient checkpointing. All training is conducted on NVIDIA H200 GPUs.

\paragraph{SFT.}
We formulate each session as a multi-turn conversation, as described in Section~\ref{sec:low_resource}. To accommodate long web observations and interaction trajectories, we segment each session using a 15-turn sliding window and truncate each example to 32K tokens. We fine-tune the model for up to three epochs with a learning rate of $2\times10^{-5}$. 

\paragraph{RL.}
We initialize RL training from the model obtained through SFT on synthetic data and optimize it using Group Reward-Decoupled Normalization Policy Optimization (GDPO)~\citep{liu_gdpo_2026}. 
We uses eight GPUs in total: four for full-parameter policy optimization with DeepSpeed ZeRO-3 and four for response generation with vLLM using tensor parallelism. 
We use a group size of 4, a global batch size of 32, and a learning rate of $1\times10^{-6}$. Based on empirical tuning, we set
$w_{\mathrm{sub}}=0.6$, $w_{\mathrm{valid}}=0.1$, and
$w_{\mathrm{exact}}=0.3$ for $R_{\mathrm{target}}$.
The RL stage takes approximately 24 hours per run.

\subsection{Performance Gap Between Synthetic and Real Supervision}
\label{app:user_modeling:SR}
Figure \ref{fig:user_modeling} shows that synthetic supervision from \framework{} improves the base model substantially but still trails supervision from real data. To understand this gap, we examined predictions from the synthetic-only model on the OPeRA test set. The residual gap concentrates in three mismatches between the source platform used for synthetic data generation and the target platform used for evaluation.

\textbf{Interaction granularity.}
The two platforms record user behavior at different levels of granularity. The source logs primarily capture actions that trigger page transitions, whereas the target platform also records interactions that modify the current view without navigating to a new page. Examples include cycling through a product image carousel, expanding a review panel, and scrolling through a long product page. Because these fine-grained interactions are absent from the synthetic corpus, the model receives no supervision for predicting them.

\textbf{Target vocabulary.}
The platforms expose different interface elements, leaving several target-domain identifiers without source-domain counterparts. For example, targets rooted at \texttt{customers\_also\_bought} or \texttt{review} correspond to recommendation and review modules that rarely appear in the source data. This mismatch further compounds the target sparsity discussed in Section~\ref{sec:low_resource}, where $77.4\%$ of unique targets occur only once. By assigning partial credit based on target similarity, RL mitigates this problem but cannot fully recover targets that receive little or no supervision.

\textbf{Operation semantics.}
Visually similar elements may also behave differently across platforms. For example, a search button opens the search interface on the source platform but submits a typed query on the target platform. The correct interaction order is therefore click-then-type in the source environment but type-then-click in the target environment. A model trained only on source-domain data learns the former ordering and transfers it to the target platform, producing action sequences that the target interface cannot execute.

The gap may narrow when synthetic interactions are generated within a simulated twin of the target interface, or within the original interface when direct agent interaction is feasible and does not expose sensitive information. Such settings would reduce the interface and functionality mismatches identified above. In our experiments, however, the original activity logs do not contain the web observations associated with each action and therefore cannot directly supervise next-action prediction. \framework{} addresses this limitation by using a powerful teacher agent to replay the logged behavioral trajectories and reconstruct multimodal web observation--action pairs. From this perspective, \framework{} serves not only as an anonymized synthetic data generation framework but also as an effective data enrichment method that transforms otherwise incomplete activity logs into training examples suitable for downstream user-modeling tasks.

\subsection{Benefits of Synthetic Data}
\label{app:user_modeling:data}

Our error analysis indicates that more than half of the exact-match improvement occurs in cases where synthetic data expands the model's effective coverage of the target space. With limited real training data, the model tends to overpredict a small set of frequent targets rather than distinguish among less common interface elements. Synthetic data exposes the model to a broader range of interactions, including suggested-item exploration and product-option selection, thereby improving its ability to identify diverse targets.

Search suggestions provide a representative example. Among the 40 test cases whose ground-truth target was a specific suggested term (e.g., \texttt{nav\_bar.suggested\_terms.bike\_bottle\_hold}), the model trained only on real data failed to predict any target correctly. Instead, it typically predicted \texttt{search\_input} or \texttt{search\_button}, two frequent targets that together account for 24.03\% of the real training data. After adding synthetic data, the model correctly identified 22 of the 40 suggested terms, achieving 55.00\% accuracy on these cases, while preserving its accuracy on \texttt{search\_input}.

We observe a similar pattern for product-option attributes. The test set contains four option types: color, scent, size, and style. Because scent does not appear in the real training data, the model trained only on real data never predicts it. Although the synthetic data also contains no scent examples, it introduces related attributes "flavor". After training with these examples, the model correctly predicts scent in some test cases, likely because flavor and scent are semantically related and occupy similar structural roles in the target schema. This result suggests that synthetic data can improve not only direct target coverage but also generalization to structurally and semantically related targets.

\subsection{Benefits of Reinforcement Learning}
\label{app:user_modeling:rl}
RL improves both output validity and target selection. Under the RL objective, which includes a format reward, the rate of malformed JSON predictions decreases from 1.99\% to 0\%. The target reward also assigns partial credit based on structural similarity, encouraging the model to identify the correct interface region even when it does not recover the exact target.

Our error analysis shows that more than half of the exact-match improvement comes from cases in which RL redirects a prediction from an unrelated top-level UI component to the correct region. For example, given the ground-truth target \texttt{buybox.purchase\_form.add\_to\_cart}, the SFT model predicts the unrelated target \texttt{reviews.popover.review\_images.next}, whereas the RL-trained model recovers the exact target. The remaining improvements primarily occur when both models identify the correct parent path but RL corrects the final leaf segment. Consistent with this pattern, after removing the final leaf segment from each target, the RL-trained model achieves 44.26\% parent-path accuracy, compared with 40.71\% for the SFT model, an improvement of 3.55 percentage points.

\end{document}